\documentclass[11pt]{article}

\usepackage{float}
\usepackage{graphicx,color}
\usepackage{epsfig}
\usepackage{amsmath,amssymb,amsthm}
\usepackage{subfigure}
\usepackage{dsfont}
\usepackage{psfrag}
\usepackage{url}

\def\1{\mathbf{1}}

\graphicspath{{figures/}}

\floatstyle{ruled}
\newfloat{Algorithm}{tbp}{lop}[section]

\begin{document}

\title{A Traveling Salesman Learns Bayesian Networks}
\author{Tuhin Sahai$^{\dagger}$ \and Stefan Klus$^{\dagger}$ \and Michael Dellnitz$^{\ddagger}$}
\date{$^\dagger$United Technologies Research Center, 411 Silver Lane, East Hartford, CT 06108, USA.\\
 $^\ddagger$University of Paderborn, Warburger Str. 100, 33098 Paderborn, Germany}

\maketitle

\begin{abstract}
Structure learning of Bayesian networks is an important problem that arises in numerous machine learning applications. In this work, we present a novel approach for learning the structure of Bayesian networks using the solution of an appropriately constructed traveling salesman problem. In our approach, one computes an optimal ordering (partially ordered set) of random variables using methods for the traveling salesman problem. This ordering significantly reduces the search space for the subsequent greedy optimization that computes the final structure of the Bayesian network. We demonstrate our approach of learning Bayesian networks on real world census and weather datasets. In both cases, we demonstrate that the approach very accurately captures dependencies between random variables. We check the accuracy of the predictions based on independent studies in both application domains.
\end{abstract}

\section{Introduction}
Bayesian networks belong to the class of probabilistic graphical models and can be represented as directed acyclic graphs (DAGs)~\cite{Cit:PGM:book}. They have been used extensively in a wide variety of applications, for instance for analysis of gene expression data~\cite{Cit:Gene}, medical diagnostics~\cite{Cit:Heckerman1992}, machine vision~\cite{Cit:Levitt:1990}, behavior of robots~\cite{Cit:Robotics2007}, and information retrieval~\cite{Cit:Info_Ret1995} to name a few.

Bayesian networks capture the joint probability distribution of the set $\chi$ of random variables (nodes in the DAG). The edges of the DAG capture the dependence structure between variables. In particular, nodes that are not connected to one another in the DAG are conditionally independent~\cite{Cit:learningtutorial}. Learning the structure of Bayesian networks is a challenging problem and has received significant attention~\cite{Cit:learningtutorial,Cit:Learning_BN:book,Cit:BNlearnsparse_1999,Cit:K2}. It is well known that given a dataset, the problem of optimally learning the associated Bayesian network structure is NP-hard~\cite{Cit:Bayesian_NP}. Several methods to learn the structure of Bayesian networks have been proposed over the years. Arguably, the most popular and successful approaches have been built around greedy optimization schemes~\cite{Cit:BNlearnsparse_1999,Cit:Chickering_learn}. Exact approaches for learning the structure of Bayesian networks have a scaling of $O(n2^{n} + n^{k+1}C(m))$, where $n$ is the number of random variables, $k$ is the maximum in-degree and $C(m)$ is a linear function of the data size $m$~\cite{Cit:ExactBN}. These approaches are based on solving a dynamic program~\cite{Cit:SinghMoore}. For large Bayesian networks the above scaling for exact algorithms is prohibitive~\cite{Cit:SinghMoore}.

In this work, we present a heuristic approach for learning the structure of Bayesian networks from data. The approach is based on computing an ordering of the random variables using the traveling salesman problem (TSP). Though using the ordering to learn Bayesian networks is not new~\cite{Cit:Ordering}, using the TSP for this task is novel. This approach provides us with the opportunity to leverage efficient implementations of TSP algorithms such as the Lin-Kernighan heuristic\footnote{LKH software~\cite{Hel98} is a popular implementation of this approach}~\cite{Cit:LKH} and cutting plane methods\footnote{Concorde TSP solver~\cite{Concorde-TSP} is an efficient implementation of a cutting plane approach coupled with other heuristics}~\cite{Cit:TSP-book} for fast structure learning of Bayesian networks.

The remainder of the paper is organized as follows. In section~\ref{Sec:learn}, we describe the approach for learning Bayesian networks using a history dependent TSP formulation. In section~\ref{Sec:Histsp}, we develop techniques for solving the history dependent TSP. We then present results on the Adult and El Ni\~{n}o datasets in section~\ref{Sec:results}. We finally draw conclusions and discuss future work in section~\ref{Sec:Conc}.

\section{Structure Learning of Bayesian Networks Using the Traveling Salesman Problem}
\label{Sec:learn}
Although we use the K2 metric~\cite{Cit:K2} to construct the Bayesian network, the only assumption our approach makes is that the scoring metric is decomposable~\cite{Cit:SinghMoore},
\begin{equation}
\mbox{GRAPHSCORE} = \displaystyle\sum_{x\in V}\mbox{NODESCORE}(x|parents(x)).
\label{eq:decomp}
\end{equation}
Thus, one can replace the K2 metric with any of the competing scoring functions such as BIC~\cite{Cit:BIC}, BDeu~\cite{Cit:BDeu}, BDe~\cite{Cit:Bde}, and minimum description length~\cite{Cit:MDL}.

A link between the optimal ordering and the TSP can be established on the basis of the decomposable metric. To find the best possible ordering $\mathbb{O}$ we start from an empty set $\phi$. We define the cost of going from $\phi$ to single random variables to be $0$. Similarly, the cost of going from any permutation of all random variables to $\phi$ is also defined to be $0$. For any partial ordering of random variables $\tilde{\mathbb{O}}$ (one that does not include all random variables) we know that,
\begin{align}
\mbox{V}(\tilde{\mathbb{O}}) = \mbox{V}(\tilde{\mathbb{O}}\setminus X) + \mbox{Cost}(X,\tilde{\mathbb{O}}\setminus X),
\label{Eq:Dyn-Prog}
\end{align}
where $X$ is a random variable, $\mbox{V}$ is the value function, $\tilde{\mathbb{O}}\setminus X$ is the set $\tilde{\mathbb{O}}$ without $X$, and $\mbox{Cost}(X,\tilde{\mathbb{O}}\setminus X)$ is the cost of adding $X$ to $\tilde{\mathbb{O}}\setminus X$.

\begin{figure}[htb]
\begin{center}
\subfigure[]{\includegraphics[width=0.5\textwidth]{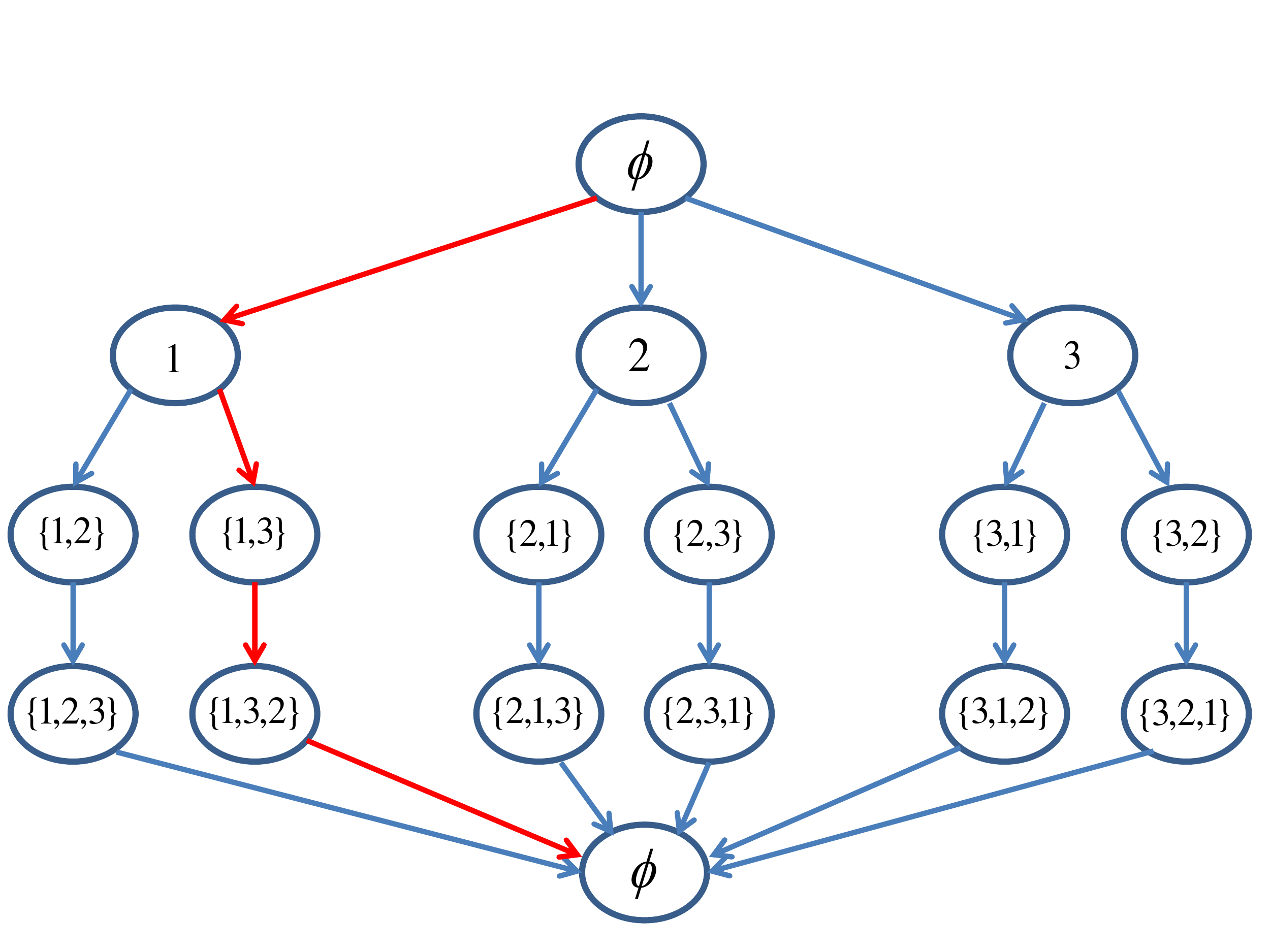}\label{Fig:DynProgBN}}
\subfigure[]{\includegraphics[width=0.3\textwidth]{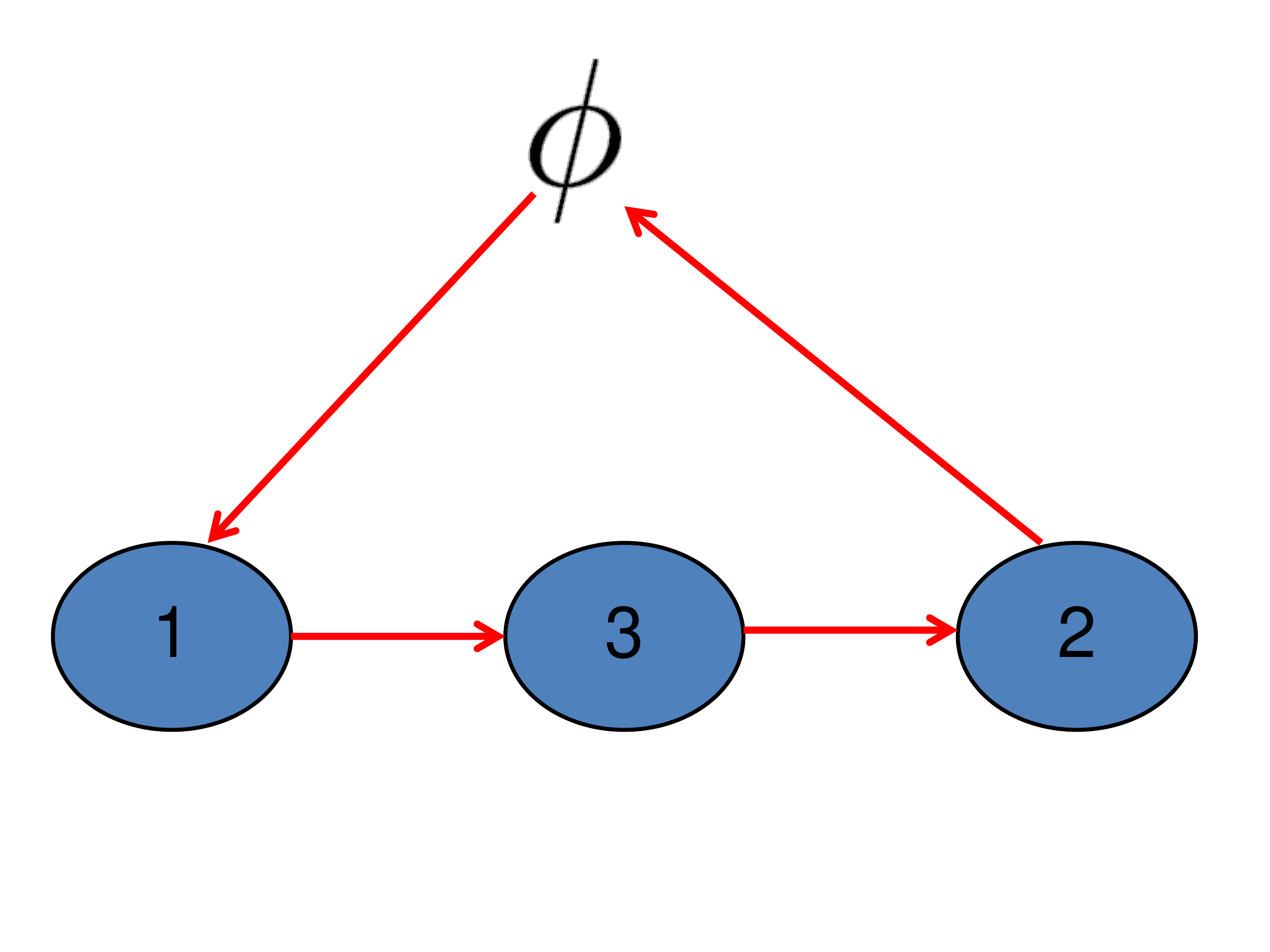}\label{Fig:TSPBN}}
\caption{a) Structure Learning of Bayesian networks as a dynamic program~\cite{Cit:SinghMoore}. The permutation tree provides the order in which nodes should be added to the list. b) The equivalent solution of the history dependent TSP for the computation of the optimal ordering.}
\label{Fig:DynamicProg}
\end{center}
\end{figure}

The above dynamic program in Eqn.~\ref{Eq:Dyn-Prog} will require $O(n^2 2^n)$ operations~\cite{Cit:SinghMoore}. Instead of solving the above equation using dynamic programming, we reformulate the problem as a history dependent TSP. This is easy to see from Eqn.~\ref{Eq:Dyn-Prog}, by considering the random variables as cities of the tour and the optimal ordering of random variables as a tour that minimizes the overall cost (see Eqn.~\ref{Eq:Hist-tsp} and Fig.~\ref{Fig:DynamicProg}).

\begin{align}
\mbox{V}(\mathbb{O}) = \mbox{min}\displaystyle\sum_{i=1}^{N}\left[\mbox{V}(\tilde{\mathbb{O}}((i+1))) - \mbox{V}(\tilde{\mathbb{O}}(i)) \right],
\label{Eq:Hist-tsp}
\end{align}

The history dependence arises due to the first term in the right hand side of Eqn.~\ref{Eq:Dyn-Prog}. The advantage of treating this minimization as a TSP, however, is the ability to leverage pre-existing TSP algorithms such as LKH~\cite{Hel98}, as discussed in the next section. Note that our approach provides Bayesian networks in which the directionality of arrows (causality) may be reversed. This may be attributed to the fact that, given the data, these networks are equally likely~\cite{Cit:PCalgo}.

\section{Solving the History Dependent Traveling Salesman Problem}
\label{Sec:Histsp}

The traveling salesman problem (TSP) is a classic problem that has received attention from the applied mathematics and computer science communities for decades. In the traditional formulation, one is given a list of city positions and tasked with finding a Hamiltonian cycle (a cycle that visits every city only once and returns to the starting city) with lowest cost~\cite{Cit:Karp}. Enumerating all possible tours becomes infeasible for problems with more than $10$ cities. In particular, the TSP is a well studied NP-hard problem~\cite{Cit:TSP-Cook}. Over several decades, many algorithms for computing the solution of the TSP have been developed; for an overview we refer the reader to~\cite{Cit:TSP-book,Cit:TSP-Cook}.

To solve the history dependent TSP, we pick Helsgaun's popular version of the Lin-Kernighan Heuristic (LKH)~\cite{Hel98}, which naturally extends to our case. LKH is a randomized approach that picks edges in the tour for removal and adds ones that are ``more likely'' to be in the optimal tour. If the replacement of edges reduces the cost, the change to the tour is accepted. The likelihood of any edge being in the optimal tour is computed using the $\alpha$-nearness that is based on minimum $1$-trees in the underlying city graph~\cite{Cit:LKH}. The LKH is the most successful approach for computing the optimal tour of TSPs with asymmetric cost~\cite{Hel98}.

\begin{figure}[htb]
\begin{center}
\subfigure[]{\includegraphics[width=0.5\textwidth]{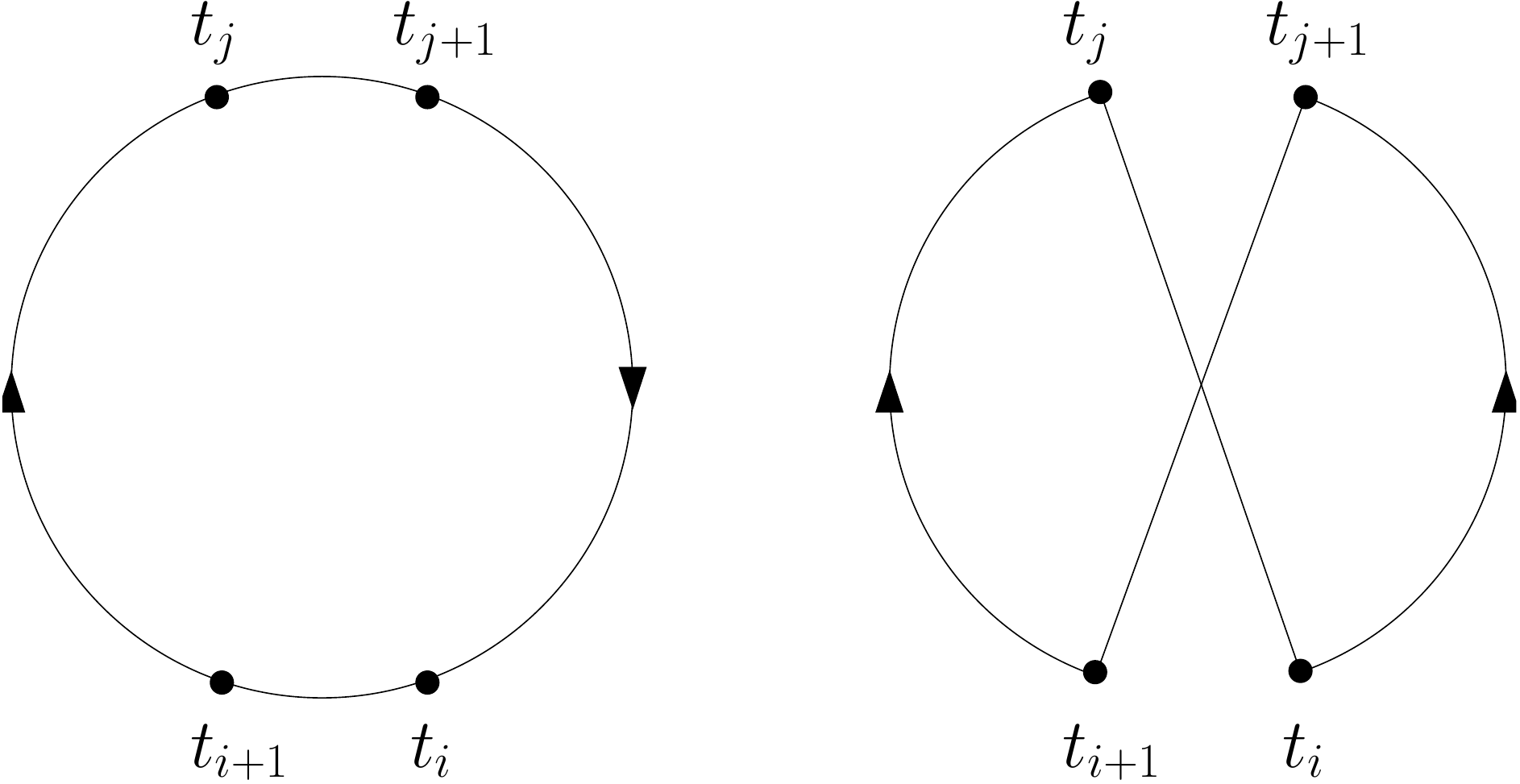}\label{Fig:2-opt}}
\subfigure[]{\includegraphics[width=0.5\textwidth]{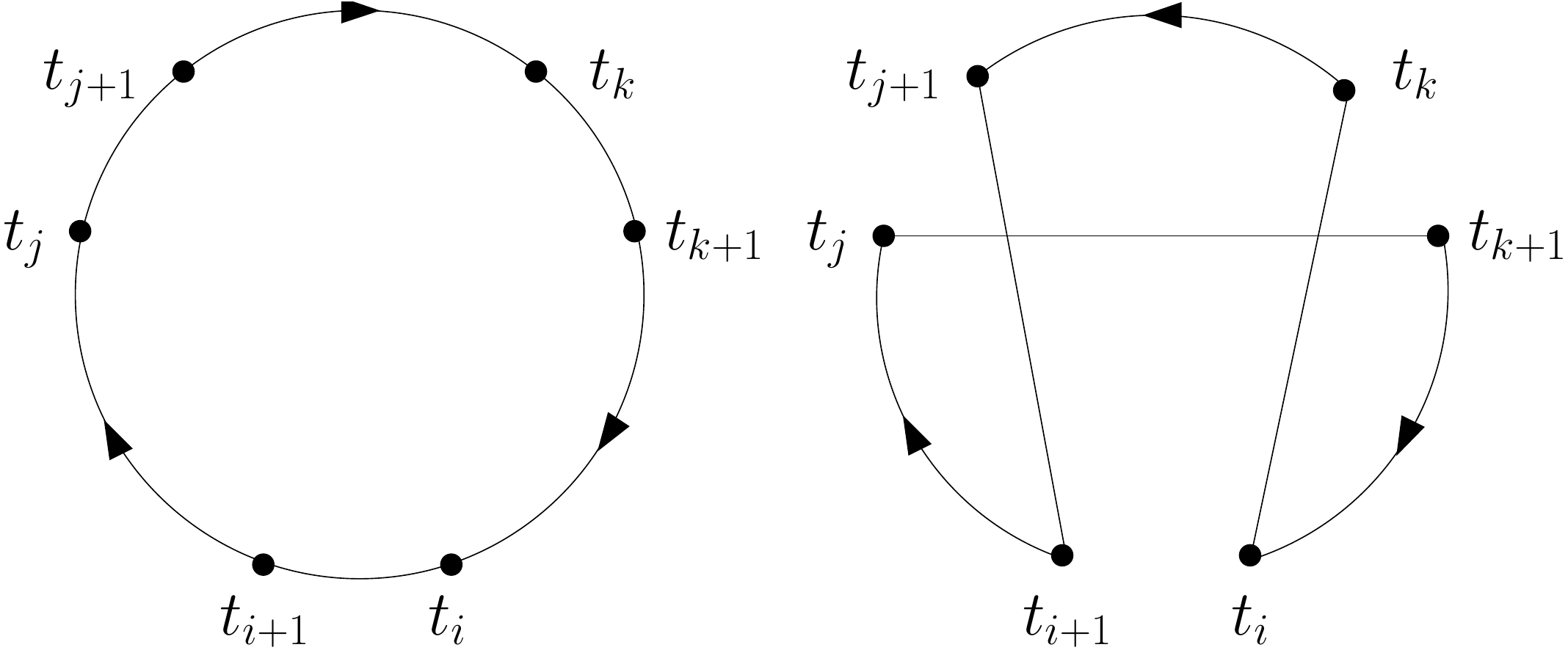}\label{Fig:3-opt}}
\caption{a) 2-opt moves for the TSP. b) 3-opt moves for the TSP.}
\label{Fig:k-opt}
\end{center}
\end{figure}

In general, one replaces $k$ edges in a simple iteration (known as $k$-opt steps). Examples of the $2$-opt and $3$-opt steps are shown in Fig.~\ref{Fig:k-opt}. Note that using higher values of $k$, in general, will give tours will lower cost. However, as $k$ increases, closing the tour becomes increasingly challenging~\cite{Hel98}.

The above approach extends naturally to the history dependent TSP. In our problem, edges again are deleted and added randomly. Unlike the standard TSP, the acceptance or rejection of the edge replacement is now dependent on the direction as well as on the existing tour. For structure learning of Bayesian networks, we compared the $2$-opt and $3$-opt iterations with Helsgaun's implementation of LKH~\cite{Hel98}. We find that despite ignoring history, the standard LKH software performs significantly better than our $2$-opt and $3$-opt implementations with history. This is, perhaps, due to the fact that LKH uses sequential $5$-opt steps as a basic move~\cite{Hel98} which is found to provide significantly better results. If Helsgaun's LKH software were to be integrated with history dependent costs, it would be expected to provide more accurate results. This is currently part of our future efforts at improving this approach. Thus, all results presented here were computed simply by using the LKH software.

\section{Results}
\label{Sec:results}
We now test our approach of computing the structure of Bayesian networks using the history dependent TSP on the Adult and El Ni\~{n}o datasets, available publicly from the UCI Machine Learning Repository~\cite{UCI-Repository}.
\subsection{Adult Dataset}
\begin{figure}%
\centerline{%
\includegraphics[width=0.7\columnwidth]{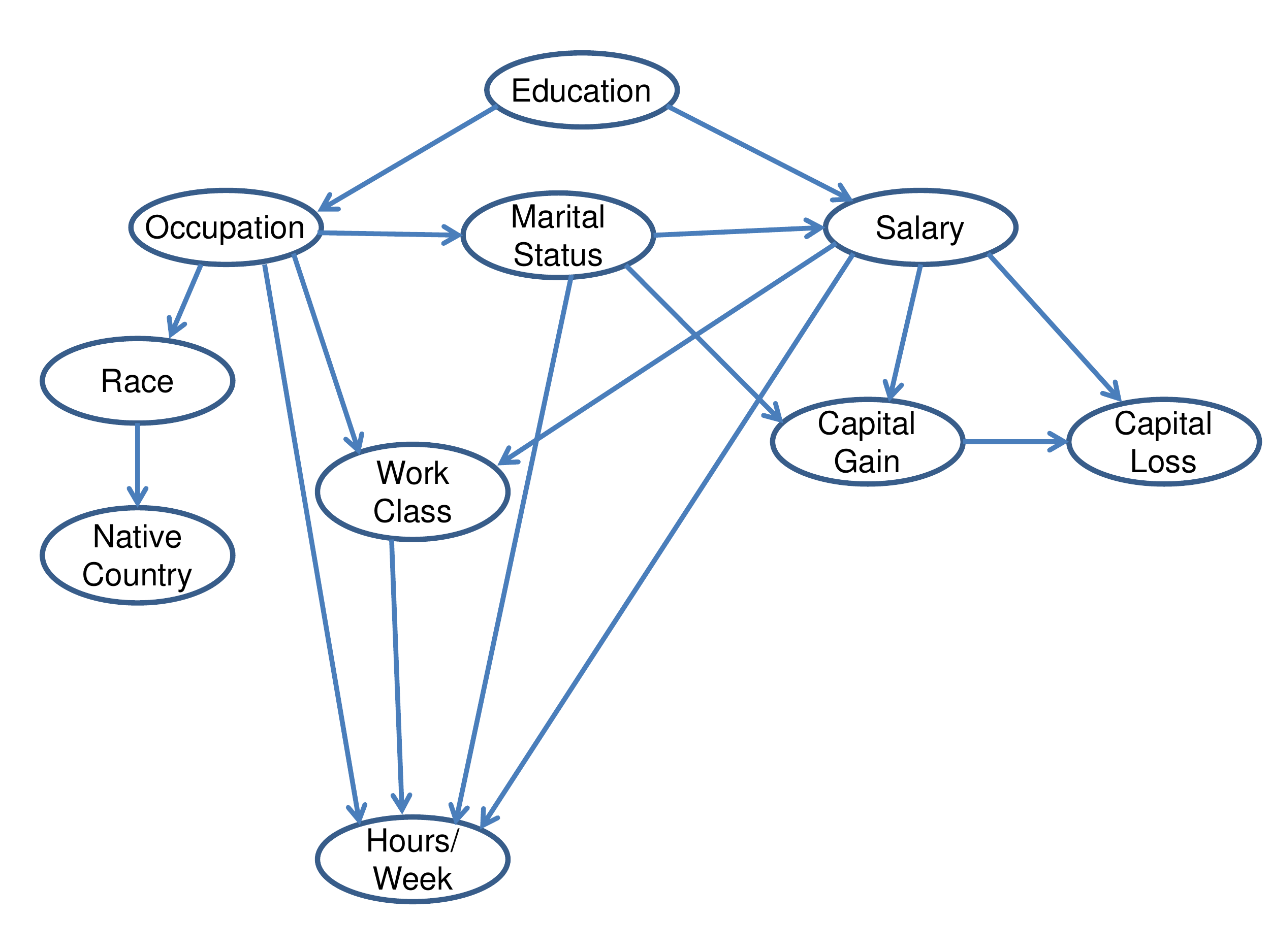}}%
\caption{Structure of the Adult dataset Bayesian network learnt using the history dependent TSP and greedy hill climbing.}%
\label{Fig:Adult-BN}%
\end{figure}

The Adult dataset was extracted from census data in $1994$ by Ronny Kohavi and Barry Becker~\cite{UCI-Repository}. The dataset consists of data for $48842$ individuals and includes several attributes including occupation, salary, number of hours worked per week, race, native country, education, marital status etc. For a complete list of attributes see~\cite{UCI-Repository}. Unfortunately, the dataset has missing values i.e. entries for certain individuals are not available. We discard these data points to finally obtain a dataset with $30162$ entries. We break this dataset into training ($29162$ entries) and testing ($1000$ entries) parts. Some of the attributes such as salary and capital gain are continuous; we discretize these attributes (for the number of possible states see table~\ref{table:adult}). We then construct a Bayesian network using our TSP and greedy hill climbing approach (shown in Fig.~\ref{Fig:Adult-BN}).

\begin{table}[h]
\centering
\begin{tabular}{|c|c||c|c|}
\hline
Work Class & 7 & Education & 16 \\
\hline
Marital Status & 7 & Occupation & 14 \\
\hline
Race & 5 & Capital Gain & 3 \\
\hline
Capital Loss & 3 & Hours/Week & 3\\
\hline
Native Country & 41 & Salary & 2\\
\hline
\end{tabular}
\caption{Number of states for each random variable in the Adult dataset. Continuous variables have been discretized.}
\label{table:adult}
\end{table}

The Bayesian network that is learnt using the TSP and hill climbing in Fig.~\ref{Fig:Adult-BN} automatically captures dependencies that are now known as a result of several independent studies. For example, the Bayesian network captures the dependency between the occupation and the number of hours worked per week~\cite{Cit:Occupation-hours-worked}. Similarly, the Bayesian network in Fig.~\ref{Fig:Adult-BN} predicts dependencies between education and salary~\cite{Cit:Education-Salary}, marital status and salary~\cite{Cit:Marital-Salary}, occupation and race~\cite{Cit:Race-Occupation}, and marital status and number of hours of worked per week~\cite{Cit:Marital-Hours}. The dependencies between race and native country, occupation and class of work, and salary and hours of work are obvious by definition and simple arithmetic respectively. Thus, we believe that the approach accurately captures the dependencies between random variables from raw data without any prior knowledge of their ordering. If one inputs an incorrect ordering of random variables, the quality of predicted dependencies degrades significantly. The comparison of resulting Bayesian networks can be performed using the log likelihood ratio~\cite{Cit:Log-Like}.

To quantitatively test the prediction of the resulting Bayesian network we check the prediction of $P(\mbox{Salary}|\mbox{Education,\,Marital Status})$. In particular, we check the accuracy of the dependence structure on predicting whether individuals in the testing dataset earn more than or less than $\$50,000$ per year (thus salary takes binary states: $0$ or $1$). We compute the mean square error using the following expression,
\begin{align}
\mbox{MSE} = \frac{1}{N_{t}} (\mathbb{E}(Y) - Y)^{2},
\label{eq:mse}
\end{align}
where $N_{t}$ is the number of data points in the testing set, $Y$ is the output state (salary in this example), and $\mathbb{E}({Y})$ is the expected value of $Y$ predicted by the Bayesian network. The MSE for the adult dataset is $0.13$. If one were to threshold probabilities at $0.5$, i.e. if the $P(\mbox{Salary}|\mbox{Education,\,Marital Status})>\$50,000$ is greater than $0.5$, we assume $\mbox{Salary}=\$50,000$. In this case we find that our approach correctly predicts the salary $78\%$ of the time.

\subsection{El Ni\~{n}o Dataset}
We now apply our algorithm to the El Ni\~{n}o dataset from the UCI Machine Learning Repository~\cite{UCI-Repository}. The data set consists of oceanographic and meteorological readings taken by buoys in the Pacific Ocean. This large dataset consists of variables such as latitude, longitude, date, zonal winds and humidity (for a complete list of variables see~\cite{UCI-Repository}). In this example, we try to answer the question of dependence of variables that was posed in~\cite{UCI-Repository}: How do the variables relate to each other?

\begin{figure}%
\centerline{%
\includegraphics[width=0.7\columnwidth]{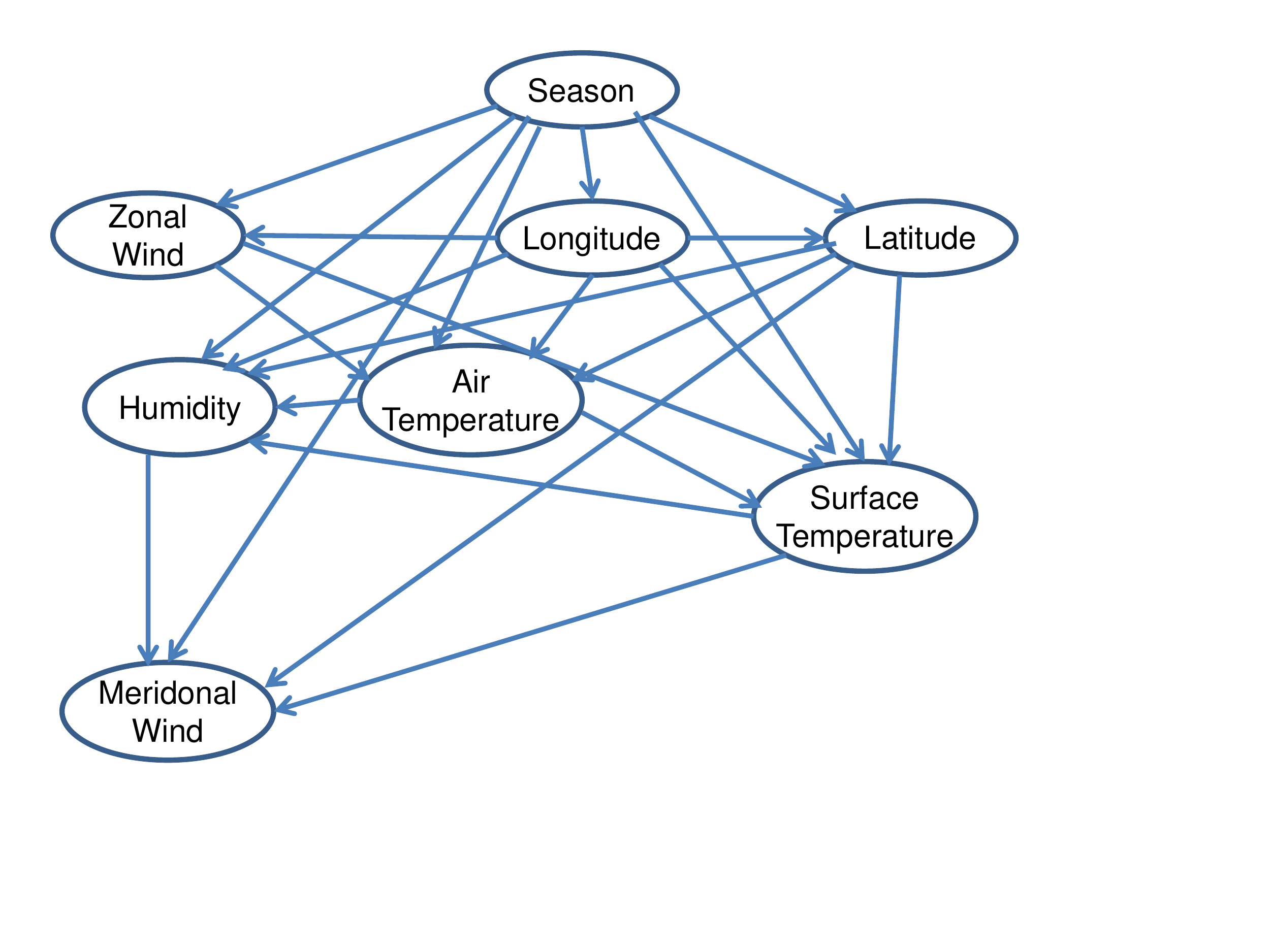}}%
\caption{Structure of the El Ni\~{n}o dataset Bayesian network learnt using the history dependent TSP and greedy hill climbing.}%
\label{Fig:ElNino-BN}%
\end{figure}

Just like in the Adult dataset example, we remove data points with missing values and partition the states into discrete values (see table~\ref{table:elnino}). After data clean up, the dataset has $93935$ data points that are used to learn the Bayesian network. We again partition the entire dataset into training (with $92935$ entries) and testing (with $1000$ entries) parts. The resulting Bayesian network is highly interconnected as seen in Fig.~\ref{Fig:ElNino-BN}. In particular, we find dependencies between air temperature and humidity, air temperature and sea surface temperature, and zonal winds and air temperature. As one would expect, we find dependencies between seasons and sea surface temperature, humidity and sea surface temperature. Note that though the predicted dependencies between seasons and longitude/latitude seem peculiar, it is to be expected since the buoys were not anchored at fixed locations and were free to drift around~\cite{UCI-Repository}. Previous analysis of this dataset considered only correlations and failed to pick up links between zonal/meridional winds and meteorological quantities. We, however, do find dependencies between the winds and meteorological quantities, suggesting a nonlinear relationship between random variables.
\begin{table}[h]
\centering
\begin{tabular}{|c|c||c|c|}
\hline
Season & 4 & Latitude & 2 \\
\hline
Longitude & 2 & Zonal Wind & 2 \\
\hline
Meridonal Wind & 2 & Humidity & 2 \\
\hline
Air Temperature & 2 & Sea Surface Temperature & 2\\
\hline
\end{tabular}
\caption{Number of states for each random variable in the El Ni\~{n}o dataset. Continuous variables have been discretized.}
\label{table:elnino}
\end{table}

To quantitatively test the predictions of the Bayesian network in Fig.~\ref{Fig:ElNino-BN}, we concentrate on predicting zonal wind speeds using seasons and longitude. Using Eqn.~\ref{eq:mse}, we find that predicted MSE is $0.09$. If we again threshold the predicted values of zonal wind at $P>0.5$, we find that the zonal wind is predicted with $89 \%$ accuracy.

\section{Conclusions and Future Work}
\label{Sec:Conc}
In this work, we have presented a new approach for learning the structure and parameters of a Bayesian network. The method computes an ordering of the random variables based on a history dependent TSP on the random variables. This ordering, typically supplied by domain knowledge experts, significantly reduces the search space for hill-climbing methods. This makes the underlying optimization techniques effective at finding Bayesian network structures that maximize likelihood.

For computing the solution of the TSP, we use the Lin-Kernighan heuristic~\cite{Cit:LKH,Hel98} with history dependent cost. The LKH approach is shown to extend naturally to this case. We used the TSP with greedy hill climbing to compute Bayesian networks to analyze the publicly available Adult and El Ni\`{n}o datasets~\cite{UCI-Repository}. We find that the approach successfully computes Bayesian networks that accurately capture the underlying system interdependencies. We check the results against common knowledge as well as domain specific studies.

Future work includes the development of novel and fast heuristics for the history dependent TSP. There is a significant lack of methods to deal with this class of problems. Additionally, to provide scalability, the authors are investigating the utility of decentralized clustering methods~\cite{Tuhin:wave} to learn Bayesian networks in distributed settings.

\section{Acknowledgements}
The authors thank Madhu Shashanka of UTRC for suggestions and discussions related to this work.

\bibliographystyle{unsrt}
\bibliography{BNStructure}
\end{document}